\documentclass[conference]{IEEEtran}
\IEEEoverridecommandlockouts
\usepackage{cite}
\usepackage{amsmath,amssymb,amsfonts}
\usepackage{algorithmic}
\usepackage{graphicx}
\usepackage{textcomp}
\usepackage{amsthm}
\usepackage{xcolor}
\usepackage{bbding}
\usepackage{makecell}
\usepackage{algorithm}  
\usepackage{color}
\usepackage{multirow}
\usepackage{bm}
\usepackage{array}
\usepackage{booktabs}
\usepackage{tabularx,booktabs}
\usepackage{multicol}
\usepackage{cleveref}
\usepackage{balance}
\usepackage{fvextra}    
\usepackage{lastpage}
\usepackage{tabulary}
\usepackage{etoolbox}
\usepackage{bbm}
\usepackage{enumitem}
\usepackage{mathrsfs}
\usepackage{multicol}
\usepackage{subcaption}
\usepackage{amsfonts,amssymb}
\usepackage{ulem}
\usepackage{cancel}
\usepackage{fancyhdr} % 添加页眉页脚
\usepackage[breakable]{tcolorbox}
\usepackage{setspace}
\usepackage{stackengine}
\usepackage{threeparttable}

\usepackage{amsthm}
\theoremstyle{definition}

\usepackage{nomencl}
\makenomenclature

\def\BibTeX{{\rm B\kern-.05em{\sc i\kern-.025em b}\kern-.08em
    T\kern-.1667em\lower.7ex\hbox{E}\kern-.125emX}}

\newcommand{\blue}[1]{\textcolor{black}{#1}}

\begin{document}

\bstctlcite{IEEEexample:BSTcontrol}

\title{Enhancing Large Language Models (LLMs) for Telecom using Dynamic Knowledge Graphs and Explainable Retrieval-Augmented Generation\\
\thanks{Dun Yuan, Hao Zhou, and Xue Liu are with the School of Computer Science, McGill University, Montreal, QC H3A 0E9, Canada. (emails:dun.yuan@mail.mcgill.ca, haozhou029@gmail.com, xue.liu@mcgill.ca); 
Hao Chen, Yan Xin, and Jianzhong (Charlie) Zhang are with Standards and Mobility Innovation Lab, Samsung Research America, Plano, Texas, TX 75023, USA. (e-mail:\{hao.chen1, yan.xin, jianzhong.z\}@samsung.com)}
}

\author{\IEEEauthorblockN{ Dun Yuan, Hao Zhou, Xue Liu, \IEEEmembership{Fellow, IEEE}, \\
Hao Chen, Yan Xin, Jianzhong (Charlie) Zhang, \IEEEmembership{Fellow, IEEE} }}

\maketitle

\thispagestyle{fancy}            %更改plain状态，首页格式设为fancy
\chead{This paper has been accepted by IEEE Wireless Communications.} 
\renewcommand{\headrulewidth}{1pt}      %把页眉线的宽度设为零，即去掉页眉线
\pagestyle{plain} 
%TC:ignore
\begin{abstract}
Large language models (LLMs) have shown strong potential across a variety of tasks, but their application in the telecom field remains challenging due to domain complexity, evolving standards, and specialized terminology. 
Therefore, general-domain LLMs may struggle to provide accurate and reliable outputs in this context, leading to increased hallucinations and reduced utility in telecom operations.
To address these limitations, this work introduces KG-RAG—a novel framework that integrates knowledge graphs (KGs) with retrieval-augmented generation (RAG) to enhance LLMs for telecom-specific tasks. 
In particular, the KG provides a structured representation of domain knowledge derived from telecom standards and technical documents, while RAG enables dynamic retrieval of relevant facts to ground the model's outputs. Such a combination improves factual accuracy, reduces hallucination, and ensures compliance with telecom specifications.
Experimental results across benchmark datasets demonstrate that KG-RAG outperforms both LLM-only and standard RAG baselines, e.g., KG-RAG achieves an average accuracy improvement of 14.3\% over RAG and 21.6\% over LLM-only models. These results highlight KG-RAG’s effectiveness in producing accurate, reliable, and explainable outputs in complex telecom scenarios.
\end{abstract}
%TC:endignore

\begin{IEEEkeywords}
Knowledge Graphs, Retrieval-Augmented Generation, Large Language Models, Telecommunications.
\end{IEEEkeywords}

%overall framework figure at intro part
\begin{figure*}[ht]
\centering
\includegraphics[width=0.95\linewidth]{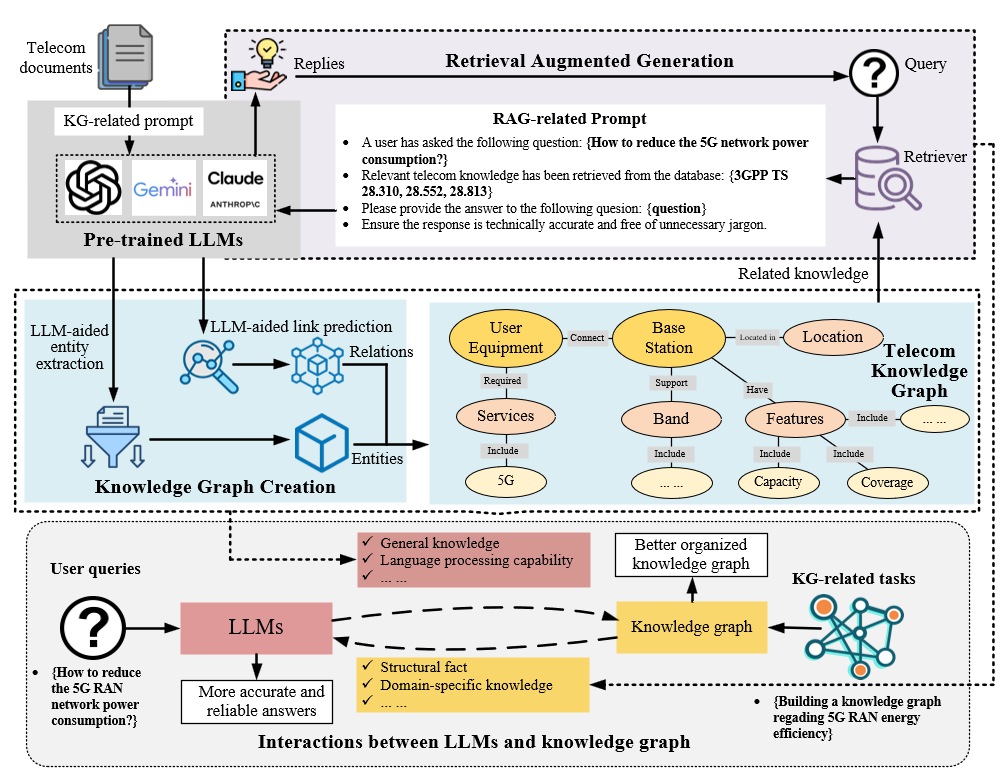}
\caption{\blue{Overview of the proposed KG-RAG framework: domain-specific telecom documents are processed by an LLM to extract entities and predict relations, which builds a KG. A retriever then surfaces relevant KG context, feeding it into an LLM prompt to generate final answers for user queries. }}
\vspace{-10pt}
\label{fig:framework}
\end{figure*}

\section{Introduction}

Large language models (LLMs) have achieved remarkable success across domains such as healthcare, education, finance, and manufacturing. They have also attracted increasing attention from the communications community~\cite{zhou2024large}, with applications explored in intent-based network management~\cite{mekrache2024intent}, power control~\cite{zhou2024large2}, traffic prediction~\cite{hu2024self}, and edge intelligence~\cite{friha2024llm}. However, most existing LLMs are designed for general-purpose use and often struggle with domain-specific tasks.
In the telecom domain, many tasks require a deep semantic understanding of network architectures, protocols, and standards, such as 3GPP~\cite{karim2023spec5g} and O-RAN specifications~\cite{nikbakht2024tspec}. Empirical results in~\cite{maatouk2023teleqna} show that GPT-4 achieves only 64.78\% accuracy on standard specifications and 74.40\% on standard overviews, highlighting the limitations of general-domain LLMs and the need for telecom-specific adaptations to improve reliability.

Several approaches have been proposed to adapt LLMs to domain-specific tasks. Fine-tuning on telecom corpora, such as technical papers and standards, can improve performance but requires substantial computational resources. Prompt engineering provides a lighter-weight alternative by guiding model outputs through carefully designed prompts~\cite{zhou2025large4}, yet its effectiveness is constrained by the inherent capabilities of LLMs and often relies on trial-and-error for complex tasks.

Motivated by these challenges, this work integrates LLMs with knowledge graphs (KGs) and retrieval-augmented generation (RAG) to enhance telecom knowledge understanding. A KG represents domain knowledge in a graph-structured form~\cite{ji2021survey}, explicitly modeling dependencies among customers, services, and network elements. For example, when answering the query ``\textit{What services will be affected if Router {F} fails?}'', a KG can directly trace downstream customers, backup paths, and service-level agreements, which is difficult for traditional unstructured knowledge systems.
RAG further improves generation by grounding LLM responses in external, verified sources such as equipment manuals, regulatory guidelines, and standards~\cite{gao2023retrieval}. It also enables access to up-to-date information, allowing responses to reflect current network states, outages, or policy changes. Compared with fine-tuning, RAG offers greater flexibility and significantly lower computational cost, making it well suited for dynamic telecom environments.

{In telecom environments, standard passage-level RAG also exhibits several important limitations. First, the retrieved evidence typically consists of long, unstructured text spans that are not aligned with domain schemas such as slices, QoS classes, or protocol stacks. Second, free-text evidence provides limited traceability: it is difficult to trace answers back to specific 3GPP releases or operational logs, even though such traceability is essential for standards compliance and troubleshooting. Third, even if the underlying corpus is periodically refreshed, a static document store may lag behind rapidly changing network configurations and KPIs, causing responses to reflect outdated system state.}
{These limitations motivate augmenting RAG with a structured, telecom-specific knowledge graph that stores ontology-aligned triples with explicit provenance and supports incremental updates. In this design, the KG acts as a front-end index and filter for RAG, narrowing retrieval to schema-consistent facts and providing a foundation for explainable, standards-compliant reasoning in downstream LLM generation.}

\begin{figure*}[t]
\centering
\includegraphics[width=0.95\linewidth]{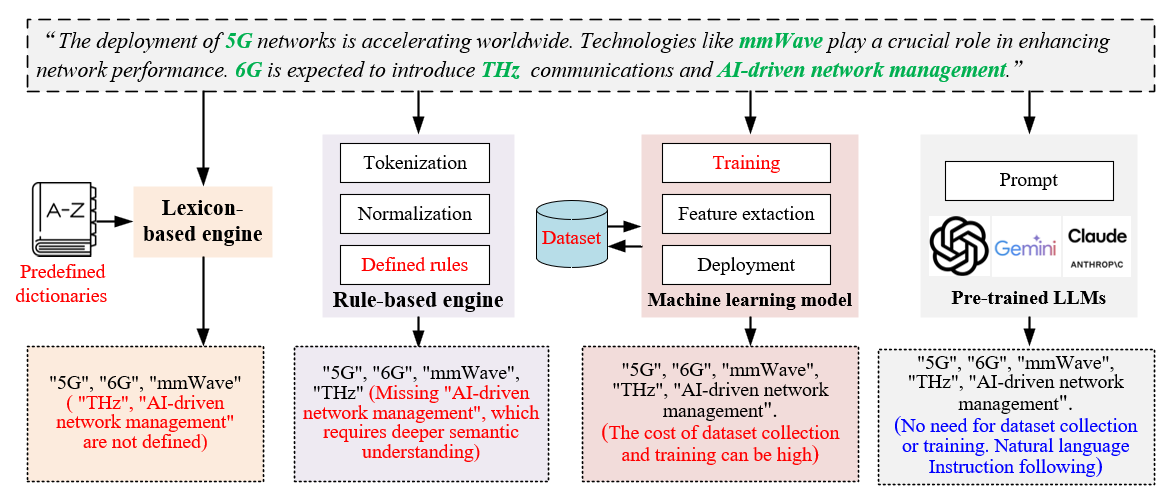}
\caption{Comparison of different entity extraction approaches: (i) Lexicon-based methods are limited to predefined dictionaries, missing newly introduced terms; (ii) Rule-based methods require explicit rules and deeper semantic understanding, limiting their adaptability; (iii) Machine learning methods effectively recognize complex terms but incur high dataset collection and training costs; (iv) Pre-trained LLMs efficiently handle emerging terms and semantic nuances without additional training.}
\label{fig:EE}
\end{figure*}

Therefore, this work enhances LLM-based telecom knowledge understanding by integrating LLMs, KGs, and RAG, as illustrated in Fig.~\ref{fig:framework}. LLMs assist KG construction through their strong language understanding and real-world knowledge, enabling accurate identification of network entities and their relationships. In turn, KGs provide structured, domain-specific facts to support LLM reasoning. For example, the telecom KG in Fig.~\ref{fig:framework} organizes radio access network knowledge across users, services, infrastructure, and features. Based on this KG, RAG retrieves relevant and up-to-date information to guide generation, improving accuracy and reducing hallucinations by grounding responses in verified knowledge sources.

Recent studies such as CommGPT~\cite{jiang2025commgpt} and the tutorial in~\cite{xiong2024tutorial} demonstrate the benefits of combining graph structures with RAG, but they primarily emphasize performance and system integration. In contrast, this work focuses on dynamic KG updates and explainable RAG. In dynamic telecom environments, where network topology, service configuration, and user behavior change rapidly, static KGs quickly become outdated. Incorporating live telemetry and configuration updates allows the KG to reflect current relationships among network functions and service flows. Moreover, KG-RAG improves explainability by verbalizing retrieved triples as atomic, schema-aligned sentences and ranking them using ontology-aware similarity, ensuring that only relevant and coherent facts are provided to the generation module.

The main contributions of this work are twofold. First, we systematically review the application of KGs in the telecom domain, covering entity extraction, link prediction, and LLM-aided KG construction, as well as RAG-enhanced LLMs for telecom tasks. Second, we propose a dynamic and explainable KG-RAG framework that constructs a graph-based telecom KG using LLMs and leverages RAG to retrieve query-relevant facts, improving accuracy, reducing hallucinations, and enhancing factual consistency in telecom scenarios.

\section{Knowledge Graphs for Telecom}

\subsection{Knowledge Graphs in Telecom Domain}
Knowledge graphs (KGs) provide a structured representation of entities and their relationships, capturing the complex interactions among network infrastructure, devices, services, and features in telecom systems. KGs model both real-world and conceptual objects as entities and explicitly describe how they are interconnected.
For instance, in the telecom domain, entities may denote network components such as base stations (BSs) and user equipment (UE), while relationships represent functional or physical connections, e.g., BS--UE channel associations and BSs providing 5G services to UEs.

To support machine learning tasks, entities and relationships are typically mapped into continuous vector spaces, enabling numerical processing. With entities and links, a KG can be visualized as a directed and labeled graph, where vertices include entities and literals, and edges represent relationships such as wireless channel connections or links between BSs and conceptual attributes like location, capacity, and coverage. In the following, we introduce two key techniques for KG creation: entity extraction and link prediction.

\subsection{Entity Extraction}
Entity extraction is a fundamental step in KG construction, aiming to identify key entities from unstructured telecom data, such as network protocols, hardware components, signal types, frequency bands, and communication standards. Several approaches exist.
Lexicon-based methods rely on predefined telecom dictionaries to match known entities, while rule-based techniques use handcrafted linguistic patterns to identify entities and relations. However, these methods may miss emerging terms when they are not explicitly defined, such as ``\textit{THz}'' and ``\textit{AI-driven network management}''.
Machine learning approaches, including supervised and unsupervised models, can generalize beyond fixed rules and lexicons but typically require annotated datasets and dedicated training. In contrast, pre-trained LLMs provide a flexible alternative for entity extraction. As illustrated in Fig. \ref{fig:EE}, documents are segmented using predefined prompts to facilitate the detection of named entities and their relationships. Trained on large corpora, LLMs can assign metadata such as network device, protocol, metric, and semantic context to extracted entities without task-specific retraining.

\subsection{Link Prediction}
Link prediction infers missing relationships between entities based on existing connections, improving the completeness and usefulness of telecom KGs. These relationships commonly involve network protocols, services, equipment, devices, and industry standards.
Translational distance models embed entities and relations into a vector space and model relationships as translation operations, while semantic matching models use bilinear transformations or tensor factorization to capture interaction patterns. Neural network-based methods further model complex, non-linear relationships; for example, convolutional approaches such as ConvE and ConvKB can identify telecom-specific patterns, including network performance bottlenecks or interoperability issues.
Recent studies show that LLM-aided methods, such as LLM-KGE and LLM-GNN, effectively enhance link prediction by leveraging semantic context and domain-specific terminology. These techniques address telecom challenges including service compatibility, network troubleshooting, and predictive maintenance, thereby enriching KG completeness and inference quality.

\subsection{Dynamic Knowledge Graph Updates}
Telecom networks are highly dynamic due to user mobility, evolving service demands, and infrastructure changes, making static KGs quickly outdated. Dynamic update mechanisms are therefore essential to reflect real-time network states. For example, Stream2Graph~\cite{barry2022stream2graph} constructs and updates KGs by ingesting heterogeneous streaming data such as network logs and telemetry, enabling applications like anomaly detection and predictive maintenance.

Dynamic KG updates support real-time network monitoring by continuously integrating telemetry from RAN elements (e.g., eNodeBs/gNodeBs), core network functions, and edge nodes, providing an up-to-date view of topology, traffic, and resource utilization. They also enable adaptive service management by reflecting radio conditions, subscriber mobility, and slice KPIs, allowing automated actions such as spectrum reallocation or scaling of mobile computing instances. Moreover, up-to-date KGs serve as a reliable basis for decision-making, supporting tasks such as capacity planning for new 5G small-cell deployments, evaluating software upgrade impacts on slice QoS, and simulating what-if scenarios based on current network relationships.

\section{Retrieval-Augmented Generation-enhanced LLM for Telecom}

\begin{figure}[!t]
    \centering
    \includegraphics[width=1.0\linewidth]{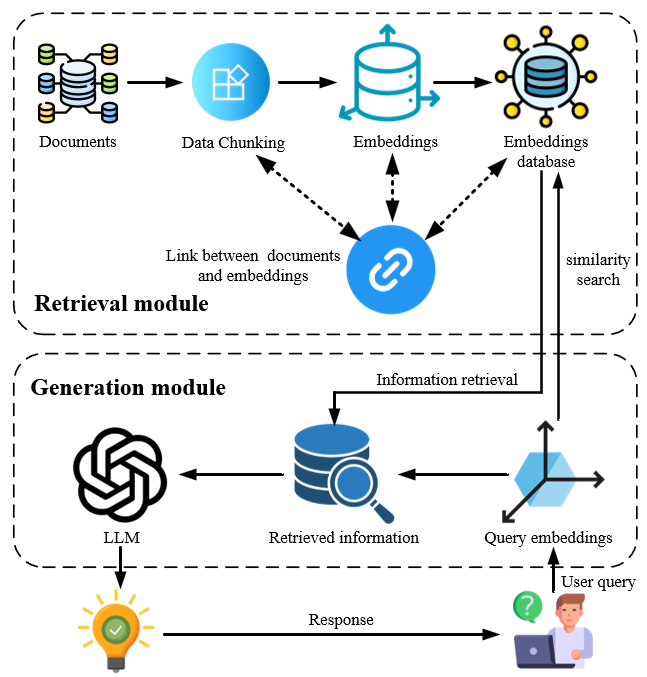}
    \caption{Illustration of a RAG workflow.}
    \label{fig:rag}
\end{figure}

LLMs can produce contextually relevant text, but purely parametric approaches can struggle with domain-specific information. 
RAG addresses such limitations by complementing LLMs with an external retrieval component. When the model encounters a query, it consults a knowledge source to gather relevant context before generating answers. 
Fig. \ref{fig:rag} presents an overview of the RAG pipeline, comprising a retrieval component that searches for relevant information, and a generation component that conditions on both the user query and the retrieved knowledge to produce a final response.

In a regular RAG workflow, the user query is first encoded into a dense vector. The external corpus is already indexed by encoding each passage into the same vector space and storing the embeddings in a nearest-neighbour index. At retrieval time, the system computes similarity scores between the query embedding and the corpus embeddings, ranking all passages and returning the top-$k$ results. These passages are concatenated and supplied to the generation module, limiting the context window to high-quality evidence and reducing hallucinations.

The retrieval process can be implemented in different approaches. For instance, cosine similarity offers a light-weight geometric metric that works well when embeddings capture rich semantics. Meanwhile, dense retrieval methods can learn task-specific embeddings and achieve higher recall on open-domain queries at the cost of heavier computation and larger indexes. In addition, hybrid schemes combine lexical and dense scores to balance precision and recall, though they add engineering complexity and require careful tuning.

After the retrieval phase identifies the most relevant knowledge entries, such as related 3GPP standards and technical papers in the telecom field, the generation component consolidates the external information with the original query to produce a contextually grounded response. 
LLMs are highly effective at generating fluent text, but their factual consistency is substantially enhanced when they can rely on clear facts.
%\subsubsection{Contextual Embedding} 
%
Specifically, as shown in Fig. \ref{fig:rag}, the generation module begins by encoding the retrieved knowledge and facts. These embeddings are then combined with the embedding of the user’s query, forming a holistic context vector. By merging these representations, the model has direct access to relevant external content, ensuring that it can reference specific details during generation.
Then, a decoder generates the response in natural language.  
Because generation is conditioned on the retrieved passages and the original query, the output is grounded in explicit evidence, lowering hallucination rates.

A key advantage of RAG is higher factual accuracy, where the model cites external knowledge instead of relying solely on its parameters. 
The above textbox shows an example of resolving a 5G slice issue.
Specifically, a telecom RAG workflow can quickly diagnose and {resolve} a 5G slice issue by: 1) retrieving a targeted subgraph such as gNodeBs with live KPIs, recent spectrum changes, a past antenna‐tilt outage, and vendor docs; 2) bundling that context into a concise prompt; and 3) generating a cause “\textit{interference shift → throughput drop; UPF queue overflow}” and three fixes, revert spectrum change, adjust antenna tilt, scale UPF. Each step is linked to its exact KG node, log entry, or document for full traceability.

\section{KG-RAG framework}

The KG-RAG framework integrates a KG with RAG to guarantee that LLM answers domain questions using standards-compliant facts rather than generic text.  Fig. \ref{fig:KG-RAG} illustrates the end-to-end pipeline: (i) The 3GPP standard document (left) is parsed into triples that populate a graph-based knowledge base (right); (ii) Then the retriever selects only the triples relevant to the user query; and (iii) The LLM receives a light, structured prompt and produces an answer that can be traced back to authoritative 3GPP sources.  In the depicted example, the framework grounds its reply on the 50\,ms latency bound and the URLLC/SST~2 mapping defined in 3GPP Release~16.

\begin{tcolorbox}[title={RAG demonstrations for resolving a 5G slice issue}]
\noindent\textbf{Situation:} OSS alerts report a sudden throughput drop and packet loss in 5G Slice 27. \vspace{5pt}

\textbf{Network Operator Query:} “What’s causing the issue in Slice 27, and how to fix it now?”\vspace{5pt}

\textbf{Retrieval:}\\  
%Subgraph of gNodeB-42/43, UPF-X/Y, MEC-1 with current KPIs; latest spectrum change at 03:15 UTC; past outage on 2025-04-12 (antenna tilt); vendor troubleshooting guide (pp. 12–15).
-\emph{Subgraph}: Identify all nodes and edges connected to Slice 27, including its gNodeBs, UPF instances, and MEC servers. gNodeB-42/43, UPF-X/Y, MEC-1 KPIs.\\
-\emph{Logs}: Fetch the last re-configuration event on those entities (e.g.\ spectrum re-allocation, VNF scaling). Latest spectrum change at 03:15 UTC.\\
-\emph{History}: Retrieve past outage reports or trouble tickets for the same cell IDs or UPF nodes. Similar outage on 2025-04-12 (antenna tilt).\\
-\emph{Docs}: Vendor’s recommended “throughput degradation” troubleshooting guide (pp. 12–15).
      
\vspace{5pt}
\textbf{Augmentation:} \vspace{-5pt}
\begin{verbatim}
Context:
- Slice 27 → gNodeB-42, gNodeB-43;...
- KPIs: gNodeB-42 throughput=450 Mbps..
- Last change: spectrum reassign on 
gNodeB-43 @ 03:15 UTC
- Past incident (2025-04-12): antenna 
tilt issue on gNodeB-42
Docs: vendor-troubleshoot.pdf (pp.15)
Question: Root cause & immediate steps?
\end{verbatim}

\textbf{Generation:} \\ 
-\emph{Cause:} Spectrum reassign altered interference → lower throughput; UPF queue overflow. \\ 
-\emph{Remedy:} Revert spectrum change (CHG-10245); tilt gNodeB-42 by –1.5°; scale UPF-X CPU +20 \%.\\  
-\emph{Provenance:} Each step tagged to its KG node/log/doc.  
\end{tcolorbox}

\subsection{Construction of the Telecom KG Database}

\begin{figure*}[t]
\centering
\includegraphics[width=1\linewidth]{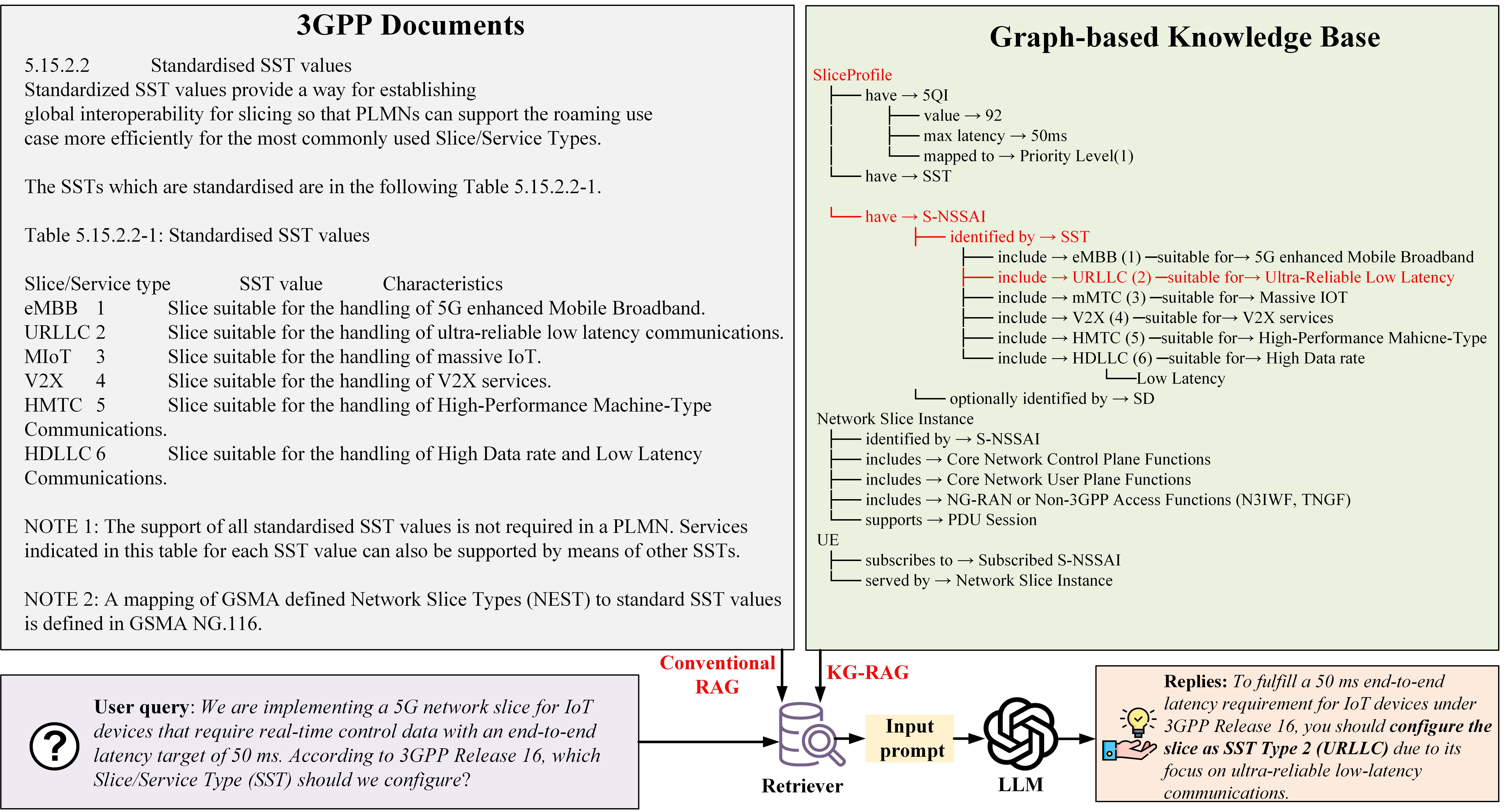}
\caption{Illustration of the KG-RAG framework applied to a 5G network use case, where an LLM retrieves from a KG created using domain knowledge from 3GPP standards. Then, it generates an answer for the input query.
}
\label{fig:KG-RAG}
\end{figure*}

\textbf{1) Document intake} We continuously ingest 3GPP technical specifications (TS~23.501, TS~23.502, TS~38.300, etc.), O\!-\!RAN Alliance working-group outputs, vendor configuration guides, and public network logs.  These documents contain formal statements such as the standardized SST table in Fig.~\ref{fig:KG-RAG} (left), where each slice/service type, such as enhanced Mobile Broadband (eMBB), Ultra-Reliable Low-Latency Communication (URLLC), massive Internet of Things (mIoT), Vehicle-to-Everything (V2X), and High-Mobility Time-Critical (HMTC) applications, is paired with an integer value.

\textbf{2) Information extraction}  A hybrid pipeline combines rule-based phrase matching with transformer-based named-entity recognition to detect domain entities (\textit{SST value}, \textit{5QI}, \textit{max latency}, \textit{Priority Level}) and relations (\textit{suitable for}, \textit{mapped to}, \textit{identified by}). 
From the sentence “\textit{Slice suitable for ultra-reliable low-latency communications}”, the system yields the triple  
\texttt{SST\_2  $\rightarrow$ suitable\_for  URLLC}.  {After LLM-based entity and relation extraction, we additionally apply a translational-distance link prediction model (TransE-style) over the partially built KG to propose missing edges, such as slice$\rightarrow$QoS-attribute and function$\rightarrow$dependency links. Each candidate edge is accepted only if (i) its relation type is permitted by the telecom ontology, (ii) its prediction score exceeds a threshold~$\tau$, and (iii) provenance metadata (source family and timestamp) can be attached. Accepted edges are stored with a flag \texttt{predicted=true} and are accessed through the same KG API as extracted edges, while explicitly extracted triples always take precedence when conflicts arise.}

\textbf{3) Schema and storage}  Triples are normalized against a domain ontology aligned with 3GPP SA2 slice-management concepts.  Nodes and relations are stored in a property graph database that supports fast traversal and attribute filtering.  All entities carry provenance metadata, such as document URI and paragraph number, so that answers remain audit-ready.

\textbf{4) Example slice profile}  The right panel in Fig.~\ref{fig:KG-RAG} shows how latency and QoS requirements attach to a \texttt{SliceProfile} node (\texttt{max latency 50\,ms}, \texttt{5QI 92}) and how that profile is linked by \textit{have\,$\rightarrow$SST} to SST~2 / URLLC. These graph edges capture the same constraints found in Release 16’s URLLC section.

\subsection{Retrieval with KG-RAG}

Given a user query, the retriever encodes it into a dense vector using a dual-encoder, where one encoder is fine-tuned specifically on telecom terminology and the other on KG triples. 
Both encoders map their respective inputs into a shared embedding space, allowing efficient similarity computation via inner product or cosine distance. 
This design ensures that the system captures not just lexical overlap but also semantic correspondence between telecom questions and KG facts.

To improve precision, retrieval is constrained by an ontology-aware filtering mechanism. Specifically, each triple is annotated with its semantic type (e.g., QoS attribute, slicing constraint, latency threshold), and queries are pre-classified using a lightweight classifier to predict their ontology type. This reduces the search space and avoids semantically unrelated triples. The top-$k$ retrieved triples are ranked based on a weighted scoring function that combines embedding similarity, ontology type match, and contextual relevance scores derived from prompt history when available.

For instance, in the case of an IoT slice provisioning query, the retriever surfaces high-confidence triples such as \texttt{SST\_2  suitable\_for  URLLC} (latency-oriented slice) and \texttt{SliceProfile  max\_latency  50\,ms} (matches the 50\,ms target).
These triples directly correspond to the query intent. Less relevant edges are filtered based on their lower semantic match and ontology mismatch. Compared with text-only RAG pipelines, which often retrieve unstructured paragraphs requiring complex interpretation, the KG-RAG approach surfaces atomic, schema-aligned triples that are easier to reason over. { During retrieval, predicted and extracted edges are ranked by the same scoring function; predicted edges never override explicitly extracted facts, and all served triples retain explicit provenance.}

\subsection{Explainable KG-RAG-based Generation}

Once the relevant triples are retrieved, they are verbalized into concise and declarative statements using a deterministic rule-based template generator. This generator maps each triple into a domain-specific sentence format, e.g., a triple like \texttt{SST\_2 suitable\_for URLLC} is rendered as “\textit{SST Type 2 is suitable for URLLC applications}.” These sentences are prepended to the user query to form the input prompt.

By structuring the prompt in this way, the model benefits from a compact and consistent context window that reflects the semantic structure of the telecom ontology. Tokens such as \texttt{S-NSSAI}, \texttt{5QI}, \texttt{SD}, and \texttt{latency\_bound} are reused verbatim from the ontology, promoting terminology consistency and reducing ambiguity. This controlled vocabulary design ensures that the LLM focuses on inference rather than extraction, narrowing its task to selecting the most plausible response given the retrieved facts.

The LLM, now conditioned on explicit, structured knowledge, generates a response such as “\textit{Configure the slice as SST Type 2 (URLLC)}.” The output is further augmented with citations to the supporting triples, and optionally, a brief explanatory clause is added, e.g., “\textit{This satisfies the 50\,ms latency requirement defined in the slice profile”}. These explanations increase interpretability and allow for step-by-step traceability, which is crucial in operational telecom contexts where AI decisions must be auditable and standards-compliant.

Explainability is a key design goal of the KG-RAG framework, particularly in the context of high-stakes telecommunications applications. Unlike conventional RAG methods that retrieve unstructured text, KG-RAG operates over a structured, ontology-aligned graph. This structure allows each retrieved fact to be represented as a discrete triple with defined semantics, enabling transparent attribution of generated responses. Every triple is linked to its provenance metadata, including the original document URI, paragraph number, and revision date. As a result, the model’s outputs can be traced line-by-line to authoritative telecom standards, facilitating auditability and enhancing trustworthiness.

This structured retrieval process supports deterministic reasoning and provides the basis for interpretable outputs. During generation, the model can optionally produce brief explanations that articulate how retrieved knowledge supports its response. For example, if multiple slice types are candidate answers, the model may clarify why a particular SST value satisfies a given latency constraint. Compared to prior RAG-based approaches that often retrieve lengthy passages requiring implicit inference, KG-RAG enables explicit and auditable reasoning steps. This enhances the reliability and transparency of AI-driven systems in telecom operations, where regulatory compliance and traceable decision-making are essential.

\begin{figure*}[t]
    \centering
    % First row
    \begin{subfigure}[t]{0.45\linewidth}
        \centering
        \includegraphics[width=\linewidth]{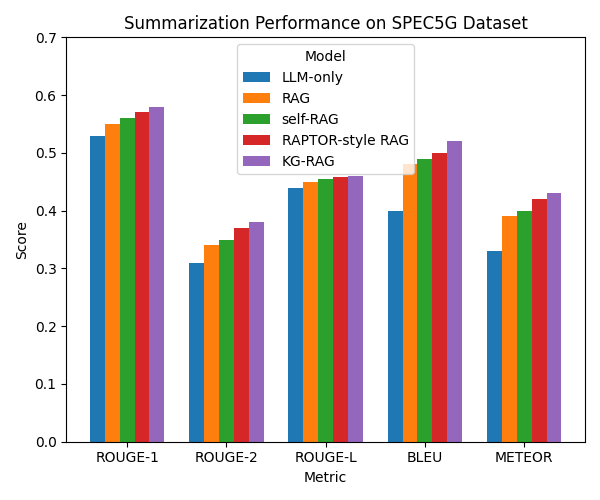}
        \caption{{Summarization metrics comparison on SPEC5G (ROUGE, BLEU, METEOR)}}
        \label{fig:rouge}
    \end{subfigure}
    \hfill
    \begin{subfigure}[t]{0.45\textwidth}
        \centering
        \includegraphics[width=\linewidth]{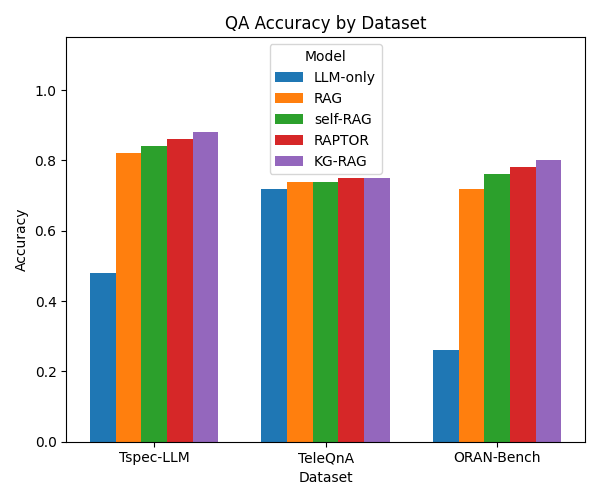}
        \caption{{QA Accuracy by Dataset (Tspec-LLM, TeleQnA, ORAN-Bench)}}
        \label{fig:qa_dataset}
    \end{subfigure}
    \vspace{0.5cm}
    % Second row
    \begin{subfigure}[t]{0.45\textwidth}
        \centering
        \includegraphics[width=\linewidth]{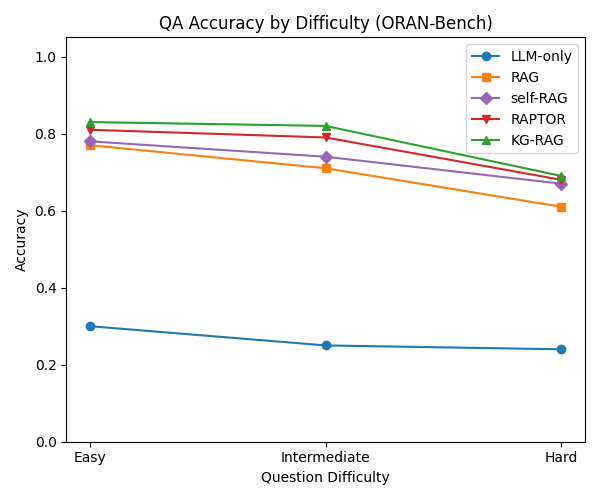}
        \caption{{QA Accuracy by Difficulty (Tspec-LLM)}}
        \label{fig:qa_difficulty}
    \end{subfigure}
    \hfill
    \begin{subfigure}[t]{0.45\textwidth}
        \centering
        \includegraphics[width=\linewidth]{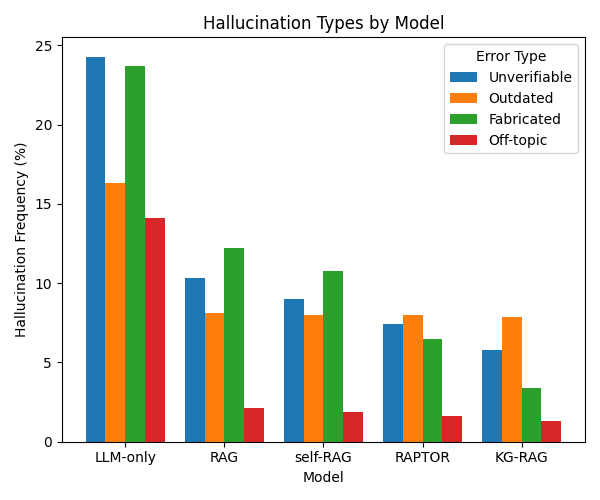}
        \caption{{Hallucination Type Breakdown (Unverifiable, Outdated, Fabricated, Off-topic)}}
        \label{fig:hallucination}
    \end{subfigure}
    \caption{Evaluation results of LLM-only, RAG, self-RAG, RAPTOR, and KG-RAG across text summarization metrics, QA accuracy, hallucination reduction, and difficulty-based robustness.}
    \label{fig:evaluation_results}
\end{figure*}

\section{Case studies}

\subsection{Experiment settings}

\blue{We utilize four benchmark datasets that reflect the diversity and complexity of real-world telecom knowledge: SPEC5G\cite{karim2023spec5g}, Tspec-LLM\cite{nikbakht2024tspec}, TeleQnA\cite{maatouk2023teleqna}, and  ORAN-Bench-13K\cite{gajjar2024oran}.
Then we conduct a series of experiments and OpenAI’s GPT-4o-mini serves as the foundational LLM in all evaluations, which is selected for its strong reasoning capabilities and efficient inference cost.}
The experiments cover two core tasks to assess the performance and applicability of the KG-RAG framework:
\textbf{1) Text Summarization}: Generating concise summaries from telecom technical documents, condensing detailed specifications into accessible and informative abstracts. 
\textbf{2) Question Answering}: Generating accurate answers to telecom-related questions by leveraging structured knowledge.

{We compare the proposed KG-RAG model against a set of representative baselines commonly used in retrieval-augmented language modeling: \textbf{1) LLM-only}: evaluating the LLM's performance without external context or structured data; \textbf{2) RAG}: using a standard passage-level RAG pipeline that retrieves document-level textual chunks from a corpus to augment the LLM’s input; \textbf{3) self-RAG}: a generator-guided RAG variant in which the LLM iteratively refines retrieval and filters inconsistent passages via self-consistency checks; and \textbf{4) RAPTOR}: a hierarchical retrieval scheme that combines lexical and dense indices with multi-scale chunking and reranking. \textbf{5) KG-RAG}: the proposed framework, which retrieves atomic, schema-aligned triples from the telecom KG instead of long text chunks, enabling more interpretable and ontology-consistent reasoning in domain-specific tasks.}

{Here we distinguish four hallucination types: unverifiable claims, where no supporting evidence can be found in the retrieved triples or cited standards/logs; fabricated facts, where statements directly contradict authoritative sources such as 3GPP or O-RAN specifications; outdated knowledge, where answers rely on content superseded by newer releases or configuration states; and off-topic content, where responses fail to address the user query.}

\subsection{KG-RAG Results Comparison and Analysis}

{
\begin{table*}[ht]
    \centering
    \caption{Overall efficiency and dynamic behavior of KG-RAG across datasets and streaming settings.}
    \begin{tabular}{c}
     \label{tab:dynamic}
            {
        \begin{tabular}{lcccc}
            \multicolumn{5}{c}{(a) Static vs Dynamic KG-RAG under streaming updates} \\[2pt]
            \hline
          Method & QA acc. (\%) & Staleness rate (\%) & Median delay (s) & P95 delay (s) \\
            \hline
            Static KG-RAG & 72.1 & 37.8 & -- & -- \\
            Dynamic KG-RAG & 84.0 & 11.4 & 5.2 & 14.7 \\
            \hline
        \end{tabular}}
        ~\vspace{10pt}
        \\
            {       
        \begin{tabular}{lccc}
            \multicolumn{4}{c}{(b) Efficiency and time cost of KG-RAG on SPEC5G, TeleQnA, and ORAN-Bench-13K} \\[2pt]
            \hline
            Metric & SPEC5G & TeleQnA & ORAN-Bench-13K \\
            \hline
            KG size (triples) & 1.2M & 0.6M & 0.4M \\
            One-time KG construction (min) & 24.6 & 13.8 & 8.9 \\
            Incremental update for 100 events (s) & 2.3 & 1.7 & 1.2 \\
            Mean KG-RAG query latency (ms) & 1080 & 860 & 920 \\
            P95 KG-RAG query latency (ms) & 1760 & 1390 & 1470 \\
            \hline
        \end{tabular}}
    \end{tabular}
\end{table*}
}

This section presents the evaluation results of the KG-RAG framework on two key tasks: text summarization and question answering, and Fig. 5 shows the results and comparisons.

Fig. \ref{fig:rouge} presents the summarization results on the SPEC5G dataset using a diverse set of metrics including ROUGE-1/2/L, BLEU, and METEOR. These metrics evaluate both lexical overlap and semantic alignment. KG-RAG achieves the highest performance across all evaluation criteria {compared with LLM-only, RAG, self-RAG, and RAPTOR methods}, indicating that retrieving structured triples helps the LLM generate more coherent, concise, and accurate summaries. {Compared with passage-level and iterative RAG variants, the structured and schema-aligned nature of the retrieved triples provides more focused contextual signals for summarization.} Notably, the gain in BLEU and METEOR suggests improvements not only in surface similarity but also in contextual fluency and meaning preservation.
Meanwhile, Fig. \ref{fig:qa_dataset} shows the QA accuracy on three telecom-focused datasets: Tspec-LLM, TeleQnA, and ORAN-Bench-13K. KG-RAG consistently outperforms the baselines{, including the optimized self-RAG and RAPTOR RAG variants}, with the most pronounced gains on ORAN-Bench, which contains standards-compliant and configuration-heavy queries. {These gains indicate that KG-RAG remains effective even when compared against stronger retrieval baselines that incorporate iterative refinement or hierarchical indexing.} Unlike RAG, which returns unstructured document chunks, KG-RAG retrieves concise knowledge triples aligned with telecom ontology, allowing the LLM to reason more effectively and reduce ambiguity in answers.

The robustness of each model under different question difficulties is evaluated in Fig. \ref{fig:qa_difficulty} {on the Tspec-LLM benchmark}. KG-RAG maintains strong performance across easy, intermediate, and hard queries, outperforming {all RAG baselines and the LLM-only methods at every level}. The most significant improvements are observed for harder questions, which require multi-hop reasoning or deep integration of technical facts. {In such cases, the explicit structure of KG-retrieved triples facilitates the aggregation of multiple related constraints and specifications.} KG-RAG’s structured and semantically filtered retrieval enables more precise factual grounding, especially in complex scenarios.
To assess factual reliability, Fig. \ref{fig:hallucination} breaks down hallucination types produced by each model. KG-RAG demonstrates the lowest hallucination rates across all categories: unverifiable claims, outdated knowledge, fabricated facts, and off-topic content. {Compared with self-RAG and RAPTOR, which still rely on unstructured textual evidence, KG-RAG further reduces hallucinations by grounding generation in ontology-aligned and provenance-aware triples.} This result highlights the framework’s ability to anchor answers in domain-grounded triples, reducing reliance on speculative or noisy generations often observed in {LLM-only and RAG-based baselines}.

\subsection{{Dynamic KG Update}}

{To demonstrate the impact of dynamic KG maintenance, we emulate a live telecom scenario by injecting a time-ordered stream of configuration and telemetry events, e.g., spectrum reassignments, gNodeB tilt adjustments, UPF scaling actions, and slice policy changes, into the system. The KG is initialized from SPEC5G, Tspec-LLM, TeleQnA, and ORAN-Bench-13K and then updated incrementally as events arrive. We compare two variants: (i) Static KG-RAG, which constructs the KG once from the corpora and keeps it frozen; and (ii) Dynamic KG-RAG, which applies the same initialization but performs incremental updates after each event based on the ontology and update pipeline described in the KG-RAG framework. For change-sensitive queries whose correct answers depend on the latest state, we report post-change QA accuracy (using the same metric as Fig.~\ref{fig:qa_dataset}), the staleness rate (fraction of answers that still reflect pre-change state), and the event-to-answer delay (time from event arrival to the first correct answer). Table~\ref{tab:dynamic} a) summarizes the results. Dynamic KG-RAG improves post-change QA accuracy from 72.1\% to 84.0\% and reduces the staleness rate from 37.8\% to 11.4\% compared with static KG-RAG, while achieving a median event-to-answer delay of 5.2\,s and a 95th-percentile delay of 14.7\,s, which is acceptable for interactive troubleshooting and operations support.}

\subsection{{Efficiency and Time Cost Analyses}}

{We further quantify the efficiency of the KG-RAG framework on SPEC5G, TeleQnA, and ORAN-Bench-13K using a fixed hardware configuration (single NVIDIA A100 GPU and a 32-core CPU). For each dataset, we measure (i) one-time KG construction time, covering document parsing and chunking, LLM-based entity and relation extraction, normalization, and graph ingestion; (ii) incremental update latency for batches of 100 configuration and telemetry events; and (iii) end-to-end KG-RAG query latency, including both retrieval and generation. The results, summarized in Table~\ref{tab:dynamic} b), show that one-time KG construction completes within tens of minutes and is incurred offline, whereas incremental updates finish within a few seconds and per-query latency remains on the order of 0.8--1.1\,s on average with 95th-percentile latency below 1.8\,s. These measurements indicate that KG-RAG can support interactive NOC analytics and engineering-support workflows while remaining consistent with the main evaluation setting.}

\section{Conclusion}
{LLMs have become a promising technology in many fields, and this work presents a novel KG-RAG framework that integrates KGs and RAG to enhance LLMs for telecom applications.}
By leveraging structured domain knowledge in telecom-specific KGs and retrieving schema-aligned facts to guide generation, our approach addresses key limitations of general-purpose LLMs, such as factual inconsistency, hallucination, and inadequate reasoning in standard-heavy tasks.
{Experiments across four telecom datasets, together with an efficiency study and a dynamic KG update case study, show that KG-RAG improves summarization and QA accuracy while maintaining interactive latency and substantially reducing stale answers on change-sensitive queries.}
Our future work will explore adaptive retrieval strategies that dynamically adjust the triple selection process based on user intent.

%Through comprehensive experiments on benchmark datasets—SPEC5G, Tspec-LLM, TeleQnA, and ORAN-Bench-13K—we demonstrate that KG-RAG consistently outperforms both standalone LLMs and traditional RAG models. Notably, it achieves superior performance in both text summarization and QA tasks, with pronounced gains in handling complex and high-difficulty queries. These improvements can be attributed to the use of atomic, ontology-grounded triples and explainable prompt structuring, which together enhance factual precision, contextual relevance, and interpretability.

%In addition, incorporating continual learning mechanisms to update KG embeddings in real time, as well as extending KG-RAG to multimodal data (e.g., diagrams, network logs, or configuration UIs), represents a promising direction for scaling telecom reasoning to broader operational scenarios. Further evaluation in low-resource and multilingual telecom settings is also a key avenue for improving accessibility and robustness.

%\section*{Acknowledgment}

\normalem
\bibliographystyle{IEEEtran}
\bibliography{Reference}

\end{document}